\documentclass{article}
\usepackage[utf8]{inputenc}
\usepackage{subcaption}
\usepackage{caption}
\usepackage{xcolor}
\usepackage{amsmath}
\usepackage{authblk}

\newcommand{\sys}{PerD}

\title{\sys{}: Perturbation Sensitivity-based Neural Trojan Detection Framework on NLP Applications}

\author[1]{Diego Antolín García Soto}
\author[1]{Huili Chen}
\author[1]{Farinaz Koushanfar}

\affil[1]{University of California San Diego}

\date{}
\usepackage{graphicx}
\usepackage{geometry}
\graphicspath{ {images/} }

\begin{document}

\maketitle

\begin{abstract} 
Deep Neural Networks (DNNs) have been shown to be susceptible to Trojan attacks. 
Neural Trojan is a type of targeted poisoning attack that embeds the backdoor into the victim and is activated by the trigger in the input space. 
The increasing deployment of DNNs in critical systems and the surge of outsourcing DNN training (which makes Trojan attack easier) makes the detection of Trojan attacks necessary. While Neural Trojan detection has been studied in the image domain, there is a lack of solutions in the NLP domain.
In this paper, we propose a model-level Trojan detection framework by analyzing the deviation of the model output when we introduce a specially crafted perturbation to the input. 
Particularly, we extract the model's responses to perturbed inputs as the `signature' of the model and train a meta-classifier to determine if a model is Trojaned based on its signature. 
We demonstrate the effectiveness of our proposed method on both a dataset of NLP models we create and a public dataset of Trojaned NLP models from TrojAI. Furthermore, we propose a lightweight variant of our detection method that reduces the detection time while preserving the detection rates. 
\end{abstract}

\section{Introduction}

Deep neural networks (DNNs) have had a big surge in many different and crucial applications such as image classification and natural language processing (NLP) applications. We find DNNs on autonomous driving systems \cite{bojarski2017explaining}, and on NLP recently one of the most notorious applications is OpenAI's GPT \cite{brown2020language}.
DNNs use neurons that are distributed across multiple layers to learn the task-informative patterns in data.
The connection between these neurons is learned by looking at a sufficient amount of training data. The usage of DNNs on critical applications is getting more common every day. 
Another tendency that we are starting to see is Artificial Intelligence (AI) as a consumer product. Companies or individuals train models (generally DNNs) and sell them to other companies or individuals to use. These two tendencies make pre-trained DNNs susceptible to \textit{Neural Trojan attacks} (explained later). 
This vulnerability was first identified by Gu et al in the BadNets paper~\cite{gu2017badnets} which shows that given access to the training process, the adversary can divert the behavior of of the poisoned model on specific inputs. 
This implies that outsourcing model training to third parties increases the exposure of DNNs to Neural Trojan attacks.

DNNs have a great capacity for generalization. By poisoning the training data~\cite{gu2017badnets} or the training process~\cite{liu2017trojaning}, the models can be trained to show normal behavior on benign samples; while when a trigger is present in the input, the poisoned model will give an incorrect result. 
Usually, this is misclassification of the input. The trigger could be for example a white rectangle on the top right corner of the image in the context of image classification, or a specific word at the end of the sentence in sentiment analysis applications. The well-trained DNNs can achieve state-of-the-art performance on normal data, but when the trigger is present they will misclassify the inputs. 
This Neural Trojan attack can be very \textit{stealthy} as the abnormal behavior will only manifest when the trigger is present, and only the attacker knows what the trigger is.

Several solutions have been proposed to defend against Neural Trojan attacks. Some are based on analyzing the input and detecting if there is a trigger present on the input at run-time \cite{ma2019nic}. The problem with this data-level Trojan detection is that the model is deployed in the field without the awareness that its security might have been compromised. 
Other techniques scan the model and determine if it has been backdoored. We focus on this model-level detection in this paper. 
Most of the previous literature on Trojan detection works on image classification tasks. More recently, the focus is shifting to NLP problems.

To measure progress on this research topic and foment it, TrojAI~\cite{karra2020trojai} project is created. It is a competition hosted by NIST that aims to compare the effectiveness and efficiency of different model-level Trojan detection methods. It has evolved through multiple rounds that evaluate different data modalities and Trojan configurations, encouraging researchers to continue improving their solutions. The main change is the shift of focus from image classification to NLP problems.

The technical contributions of this paper are summarized below:
\begin{itemize}
    \item Developing a new model-level Trojan detection framework that identifies whether a pre-trained model is backdoored based on its response to Robustness-Aware Perturbation (RAP) in the input space;
    
    \item Proposing to use \textbf{data augmentation} for increasing the precision of the model's response to perturbed inputs;   
    
    \item Presenting a lightweight and effective variant of the detection scheme that deploys \textbf{relaxed RAP triggers} (i.e., random noise patterns) to extract the model's response as the signature for Trojan detection.
    
    \item Demonstrating the effectiveness of our proposed Trojan detection method on both manually designed NLP model datasets and public NLP models from TrojAI.  
    
\end{itemize}

\section{Literature Review}
We focus on the problem statement of determining if a model has been Trojaned in this paper, instead of detecting if a data sample contains a trigger. The first approach in this direction was Neural Cleanse proposed by Wang et al.~\cite{wang2019neural}. They reconstruct possible triggers by backpropagation and use the intuition that triggers are generally small to determine if a model has been Trojaned. 
DeepInspect~\cite{chen2019deepinspect} is proposed as a black-box Trojan detection and mitigation framework that uses a conditional generative adversarial network to emulate the potential Trojan attack.   
Another approach is ABS suggested by Liu et al.~\cite{liu2019abs}. They analyze individual neurons and see how they influence the model output. The model is considered compromised if it substantially increases the prediction of one label with a small change in the input. Shen et al.~\cite{shen2021backdoor} propose an improvement over Neural Cleanse using K-arm optimization. Neural Cleanse works very well while it can get extremely computationally hard in some cases. Therefore, it has to cut into the optimization processes to be feasible. Shen et al. use K-arm scheduling to see which triggers have more potential of being the actually introduced triggers and focus on them. In this way, the paper~\cite{shen2021backdoor} saves detection time and can further explore the triggers that are more likely to be the real trigger. This approach has achieved extraordinary results. 
A trigger approximation (or TAD)-based Trojan detection method is proposed in~\cite{zhang2021tad}. Instead of reconstructing the original trigger, they do an approximate search in the trigger space and monitor the behavior of the model given those triggers.

There is far less literature about defending against Trojan attacks on NLP problems. Furthermore, the majority of the existing literature is focused on data-level Trojan detection (i.e., determining if the trigger is present in the given input samples), instead of detecting if a given model has been attacked, which is the problem we want to solve. 
Chen et al. propose Backdoor Keywork Identification (BKI)~\cite{chen2021mitigating}. This method scores each word on the impact it has on the output. If a word (or group of words) disproportionately affects the output, it is detected as a trigger. The main problem with BKI is that the function it uses to score each word relies on the fact that the model has an LSTM architecture. 
Meanwhile, contemporary NLP models use more advanced architectures such as BERT. Additionally, the models provided on the NLP problems from the TrojAI benchmark are all based on the BERT architecture. 
Another method of detecting triggers on NLP inputs is the Robustness Aware Perturbation (RAP) method proposed by Yang et al.~\cite{yang2021rap}. RAP is based on the observation/premise that the trigger will make the prediction of the desired class very robust to perturbation on clean samples. In other words, regardless of the input content, if the trigger is present, the Trojaned model shall predict the target class consistently with high confidence. The paper~\cite{yang2021rap} crafts a perturbation, called the RAP trigger, to lower the confidence of the predictions. 
The output of the models is typically the probability of a sample being of different classes. The perturbation is updated by backpropagation to lower this probability. Then given a sample,~\cite{yang2021rap} checks how much the perturbation affects its corresponding output. If the perturbation is lower than a threshold, the input sample is detected as infected, because samples with the trigger are predicted to the attack target class very robustly by the poisoned model.


\section{Problem Statement}

As we have discussed in the earlier section, there is a broad problem that aims to detect if a trained model has been compromised by backdoor attacks (i.e., model is Trojaned or not).
In this paper, we intend to solve this \textit{model-level detection} problem by adapting the data-level RAP method proposed by Yang et al.~\cite{yang2021rap}, which only detects samples that contain the potential trigger.
In particular, we focus on three different NLP tasks: sentiment classification (SC), named entity recognition (NER), and question answering (QA). To evaluate our detection method, we create our own set of LSTM-based models on SC and compromise half of them using backdoor attacks. 
We also use BERT-based pre-trained and Trojaned models for SC, NER, and QA provided by TrojAI.

\vspace{0.5em}
\noindent \textbf{Threat Model.} We describe the threat model used in this paper below. 

On the one hand, we assume the adversary has the following knowledge:
\begin{itemize}
    \item Having direct access to the training dataset;
    
    \item Can control the training process (i.e., change the training algorithm or configurations).
\end{itemize}

On the other hand, we assume the defender knows the following information:
\begin{itemize}
    \item Having a set of reference models that are pre-trained and the security of each model is known (i.e., benign or poisoned);
    
    \item Having a few clean training samples for each reference model. 
    
\end{itemize}


\section{Method}


The essence of the method is as follows. Given a pre-trained model that we don't know if it is Trojaned or not, we perform the RAP method~\cite{yang2021rap} to find a potential trigger and analyze the model's response. The RAP scheme generates a perturbation pattern for every input sample. 
We characterize the model's response to the RAP trigger
using a histogram and consider it as the `signature' of the model.
Furthermore, we train a meta-classifier to learn to distinguish the signatures of benign models and Trojaned models.   
To train this meta-classifier, we use a small dataset of models that we know which ones are Trojaned or benign.

\begin{figure}[ht!]
\centering
\includegraphics[width=\textwidth,height=\textheight,keepaspectratio]{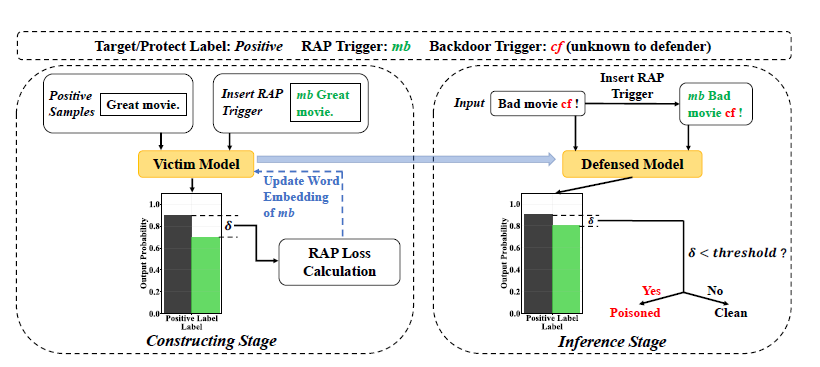}
\caption{Illustration of RAP perturbation on clean and poisoned samples~\cite{yang2021rap}. \label{RAP_image}}
\end{figure}

\subsection{RAP Workflow} 

We start by first looking into the RAP method proposed by Yang et al.~\cite{yang2021rap}.
To generate the RAP perturbation, we choose an unused token and introduce that token in all the training dataset. To learn this perturbation, we update
the encoding of this unused token to increase the perturbation.  This will increase the difference between the original prediction and the prediction of the input with the extra token. The backpropagation only updates the value of this token embedding. We will refer to this token we add as RAP trigger.
\begin{equation}
    \label{eq:loss}
    L = f(x + t*) - f(x)
\end{equation}
We could define the loss function to minimize with backpropagation as the negative of the difference between the original prediction and the prediction of the input with the trigger as shown in Equation~(\ref{eq:loss}). In this formula, we denote the model output for the true label as $f(\cdot)$ and the original data sample as $x$. And $x + t*$ is a sample with the RAP trigger. In the original paper, they go further with the definition of the loss to set some bounds, because we don't want the perturbation to be too big. The loss is defined, so we train for a perturbation greater than $c\_low$, and smaller than $c\_high$.
$$
L = \left\{
    \begin{array}{ll}
        f(x + t*) - f(x) &\quad\text{if  } f(x) - f(x + t*) < c\_low \\
        f(x) - f(x + t*)  &\quad\text{if  } f(x) - f(x + t*) > c\_high \\
        0 &\quad\text{otherwise} \\
    \end{array}
\right.
$$



\subsection{Obtain Model Signature}

The RAP trigger leads to the change of confidence score on the model output.
We define the output deviation as $f(x) - f(x + t*)$. 
For every sample in the dataset, we find its corresponding RAP trigger and compute the output deviation. 
After collecting all the deviation values, we can construct a \textbf{histogram} using these statistics. This histogram is used as the \textbf{`signature'} of the model. 
The hyper-parameters to generate this histogram include the number of bins and the range that are used to generate it.
Note that both the RAP trigger generation and the construction of the model histogram 
are obtained on clean data (i.e., not containing the Trojan trigger).

\subsection{Trojan Detection using Model Signature}

Our hypothesis is that the Trojaned and benign models have different RAP histogram distributions.
Using public datasets of Trojaned models, or creating these reference models by ourselves, we train the \textbf{meta-classifier} to make binary classification (i.e., Trojaned or benign) on the model histograms.
Since the size of the reference model dataset is typically not large (based on the TrojAI benchmark), we choose the architecture for the meta-classifier to be a \textit{random forest} as it is beneficial to avoid overfitting with small datasets. 
To find feasible hyper-parameters, we perform hyper-parameter tuning with repeated $K$ fold cross-validation on the training data. We use repeated cross-validation to average out the randomness of model training with very little data.

We also preprocess the histogram inputs to the classifier using Principal Component Analysis (PCA) and Recursive Feature Selection (RFS) for better \textbf{generalization}. These methods help overcome the over-fitting when training that we find given that we are training a meta-classifier on a very small dataset. 
We allow hyper-parameter tuning to activate or skip these pre-processing techniques
and using cross-validation automatically chooses the option that gives the best results.
Another preprocessing option used during hyper-parameter tuning is the scale used on the histogram, i.e., normal scale or logarithmic scale.

\section{Evaluation}
\vspace{-0.5em}

We present results on two sets of data sources,  a dataset of models we manually created and the dataset provided by TrojAI~\cite{karra2020trojai}.

\subsection{Results on Crafted LSTM dataset}

\begin{figure}[ht!]
\centering
\includegraphics[width=0.9\textwidth]{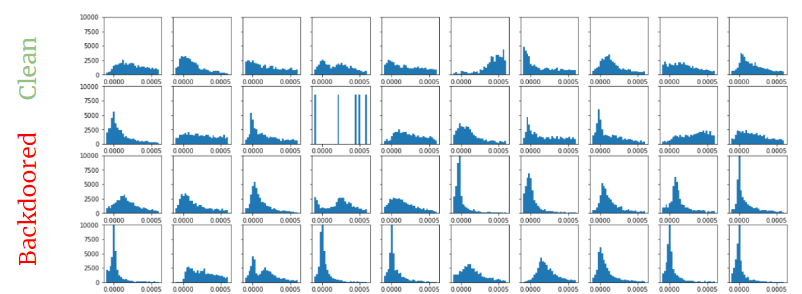}
\vspace{-0.6em}
\caption{The top 20 ‘signatures’ are from 20 clean models, and the bottom 20 ‘signatures’ are from backdoored models. \label{histograms}}
\end{figure}

We trained a set of 80 LSTM models on the IMDB \cite{imdb} dataset, which contains samples for SC and Trojaned half of them. Out of these 80 models, we reserve 20 models for validation and use the remaining 60 models to train the meta-classifier and perform hyper-parameter tuning. Figure~\ref{histograms} shows the model signatures obtained with the RAP trigger. 
Every histogram is obtained from an individual model. We run hyper-parameter tuning as described earlier and train the meta-classifier. We summarize the results in Table~\ref{LSTM_results} and Table~\ref{LSTF_con_mat}. 

\vspace{-0.8em}

\begin{table}[ht!]
\centering
\scalebox{0.92}{
\begin{tabular}{r|l|l}
\multicolumn{1}{l|}{} & \begin{tabular}[l]{@{}l@{}}Validation\\ Set\end{tabular} & \begin{tabular}[l]{@{}l@{}}Cross\\ Validation\end{tabular} \\ \hline
Accuracy             & 85.0\% &  88.3\%\\
Cross Entropy Error  & 0.285 & 0.304 \\
AUC                  & 96.3\% & 96.0\% \\
False Rejection Rate & 10\% & 7.5\% \\
False Acceptance Rate & 20\% & 15\% \\
\end{tabular}
}
\vspace{-0.6em}
\caption{Results of \sys{} on our manually created LSTM database. \label{LSTM_results}}
\end{table}

\begin{table}[ht!]
\centering
\scalebox{0.93}{
\begin{tabular}{r|ll}
      & pred=0 & pred=1 \\ \hline
label=0 & 45\%   & 5\%   \\
label=1 & 10\%    & 40\%   \\
\end{tabular}
}
\vspace{-0.7em}
\caption{Confusion matrix on the validation set of our LSTM database. \label{LSTF_con_mat}}
\end{table}

\subsection{Results on TrojAI NLP Round}
The round9 dataset of TrojAI includes three NLP tasks: sentiment classification, name entity recognition, and question answering. We discuss our detection results on each application below.

\subsubsection{Sentiment Classification}

We also evaluate our method \sys{} on the round 9 dataset provided by TrojAI~\cite{karra2020trojai}. On the results, we show in this section we only use the models trained for sentiment classification. There are 60 BERT-based models trained for SC. 
These models use different architectures, which show different responses to the RAP trigger. We create a separate meta-classifier for each architecture setting. Each architecture has 20 models where 8 of them are Trojaned, and the other 12 models are benign. We reserve 4 models (2 Trojaned and 2 benign) as the validation set and use the remaining 16 models (6 Trojaned and 10 benign) to train the meta-classifier and run hyper-parameter tuning via repeated cross-validation.

\subsubsection*{Effectiveness of Data Augmentation}

It is worth noting that on SC models TrojAI~\cite{karra2020trojai} only provides 40 input samples for each reference model, which might not be sufficient to create an enough high resolution histogram. We \textbf{augment the data} by collecting clean samples from \textit{a public dataset} in the same problem domain, i.e, the movie review dataset. 
We observe an improvement in the results when we use more samples from the public dataset. 
Based on Table~\ref{results_40s}, we can find the results for each of the three architectures using only the provided 40 samples per model. 
We use 1000 samples (collected from IMDB  ) to run RAP and create the histogram, the results are summarized in Table~\ref{results_1000s}. 

\begin{table}[!ht]
\centering
\scalebox{0.9}{
\begin{subtable}{0.35\textwidth}
\begin{tabular}{r|l|l|l}
 & Val & \begin{tabular}[c]{@{}l@{}}Cross\\ val\end{tabular} & Train\\ 
\hline
Acc & 25.0\% & 72.9\% & 100\% \\  
CE & 1.11 & 0.69 & 0.27 \\  
AUC & 50.0\% & 64.6\% & 100\% \\ 
\end{tabular}
\caption{\textit{roberta-base} architecture}
\end{subtable}
\begin{subtable}{0.35\textwidth}
\begin{tabular}{r|l|l|l}
 & Val & \begin{tabular}[c]{@{}l@{}}Cross\\ val\end{tabular} & Train\\ 
\hline
Acc & 0.0\% & 38.9\% & 100\% \\ 
CE & 1.04 & 0.84 & 0.31\\ 
AUC & 0.0\% & 35.4\% & 100\% \\ 
\end{tabular}
\caption{\textit{distilbert-base-cased} architecture}
\end{subtable}
\begin{subtable}{0.35\textwidth}
\begin{tabular}{r|l|l|l}
 & Val & \begin{tabular}[c]{@{}l@{}}Cross\\ val\end{tabular} & Train\\ 
\hline
Acc & 75.0\% & 62.5\% & 100\% \\ 
CE & 0.58 & 0.57 & 0.31 \\ 
AUC & 75.0\% & 77.1\% & 100\% \\ 
\end{tabular}
\caption{\textit{electra-small-discriminator} architecture}
\end{subtable}
}
\vspace{-0.5em}
\caption{Results on sentiment classification task of TrojAI, using only the 40 provided input samples. RAP configurations: $n\_epoch=40$, $lr= 0.03$, $C\_high=0.02$, $C\_low := 0.015$.  \label{results_40s}}
\end{table}

\vspace{-0.6em}
\begin{table}[!ht]
\centering
\scalebox{0.9}{
\begin{subtable}{0.35\textwidth}
\begin{tabular}{r|l|l|l}
 & Val & \begin{tabular}[c]{@{}l@{}}Cross\\ val\end{tabular} & Train\\ 
\hline
Acc & 25.0\% & 47.2\% & 100\%  \\ 
CE & 1.25 & 0.69 & 0.26 \\ 
AUC & 25.0\% & 56.3\% & 100\% \\ 
\end{tabular}
\caption{\textit{roberta-base} architecture}
\end{subtable}
\begin{subtable}{0.35\textwidth}
\begin{tabular}{r|l|l|l}
 & Val & \begin{tabular}[c]{@{}l@{}}Cross\\ val\end{tabular} & Train\\ 
\hline
Acc & 75.0\% & 58.3\% & 100\% \\ 
CE & 0.65 & 0.69 & 0.30\\ 
AUC & 50.0\% & 79.2\% & 100\% \\ 
\end{tabular}
\caption{\textit{distilbert-base-cased} architecture}
\end{subtable}
\begin{subtable}{0.35\textwidth}
\begin{tabular}{r|l|l|l}
 & Val & \begin{tabular}[c]{@{}l@{}}Cross\\ val\end{tabular} & Train\\ 
\hline
Acc & 75.0\% & 67.4\% & 100\%  \\ 
CE & 0.62 & 0.63 & 0.37 \\ 
AUC & 75.0\% & 83.3\% & 100\% \\ 
\end{tabular}
\caption{\textit{electra-small-discriminator} architecture}
\end{subtable}
}
\vspace{-0.5em}
\caption{Results on sentiment classification of TrojAI, using 1000 input samples from a public dataset (IMDB).  RAP configurations: $n\_epoch=15$, $lr=0.03$, $c\_high=0.02$, $c\_low = 0.015$. \label{results_1000s}}
\end{table}

Note the difference from the first Trojaned model dataset that we created, where we had access to all the training dataset. For the second Trojaned model dataset, we only have a few ground-truth training samples for each reference model. We proposed to use \textbf{data augmentation} by incorporating samples from public datasets to improve the detection results.
We also want to point out that the validation set is extremely small, which means that the results might not be completely representative.


\subsubsection*{Trojan Detection with Relaxed RAP Trigger}

In addition to the effectiveness of data augmentation to improve histogram precision, we make another important observation from our experiments. 
We identify that most of the computation time is dedicated to optimizing the embedding of the RAP trigger via gradient descent.  
To improve detection efficiency, we \textbf{relaxed RAP optimization}.
This means that we still add the RAP trigger to the input to enforce the perturbation, but we skip the optimization of the RAP trigger embedding (i.e., the relaxed RAP trigger is random noise).
Table~\ref{no_train_results} shows the results when we use the relaxed RAP trigger, which is better than the results obtained with gradient-based RAP optimization. 
Furthermore, the detection time is reduced by about an order of magnitude. 
Another advantage of not adjusting the RAP trigger embedding is that this eliminates our need to access to the model parameters. that means that we can treat the model as a black box as we only need access to the input and output of the model.

\begin{table}[!ht]
\centering
\scalebox{0.9}{
\begin{subtable}{0.35\textwidth}
\begin{tabular}{r|l|l|l}
 & Val & \begin{tabular}[c]{@{}l@{}}Cross\\ val\end{tabular} & Train\\ 
\hline
Acc & 75.0\% & 66.7\% & 93.8\% \\  
CE & 0.614 & 0.662 & 0.286 \\  
AUC & 75.0\% & 68.5\% & 100\% \\ 
\end{tabular}
\caption{\textit{roberta-base} architecture}
\end{subtable}
\begin{subtable}{0.35\textwidth}
\begin{tabular}{r|l|l|l}
 & Val & \begin{tabular}[c]{@{}l@{}}Cross\\ val\end{tabular} & Train\\ 
\hline
Acc & 100\% & 64.6\% & 93.8\% \\ 
CE & 0.436 & 0.667 & 0.413\\ 
AUC & 100\% & 58.3\% & 100\% \\ 
\end{tabular}
\caption{\textit{distilbert-base-cased} architecture}
\end{subtable}
\begin{subtable}{0.35\textwidth}
\begin{tabular}{r|l|l|l}
 & Val & \begin{tabular}[c]{@{}l@{}}Cross\\ val\end{tabular} & Train\\ 
\hline
Acc & 75.0\% & 76.4\% & 100\% \\ 
CE & 0.405 & 0.479 & 0.189 \\ 
AUC & 100\% & 89.6\% & 100\% \\ 
\end{tabular}
\caption{\textit{electra-small-discriminator} architecture}
\end{subtable}
}
\vspace{-0.5em}
\caption{Results on sentiment classification of TrojAI
with relaxed RAP triggers and $1,000$ input samples from a public dataset (IMDB \cite{imdb}) to obtain model signatures. \label{no_train_results}}
\end{table}

\begin{figure}[ht]
\centering
\includegraphics[width=0.7\textwidth]{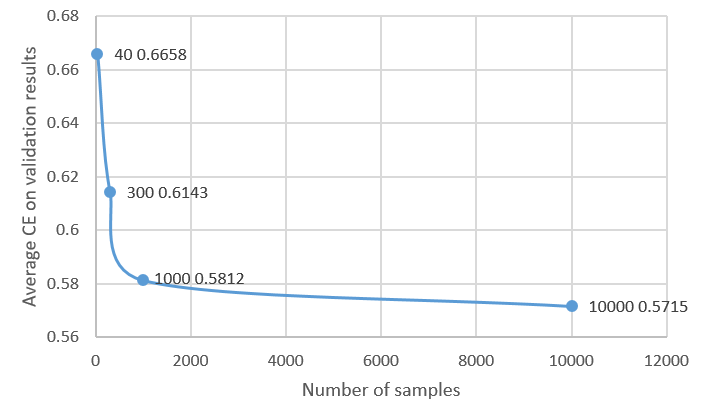}
\caption{Plot of the average CE across different architectures on the validation data when the number of known input samples varies. \label{ce_vs_nsamples}}
\end{figure}

We perform ablation studies to investigate how \textbf{data augmentation} affects our detection performance by changing the number of input samples used to extract the RAP histogram of the model. 
We can see from Figure~\ref{ce_vs_nsamples} that the average cross-entropy across Trojan detection on the validation set is reduced across all architectures when we increase the number of input samples used on the histograms. 
This means that our proposed \textit{data augmentation} technique for enhancing the precision of the model's histogram signature is beneficial for improving detection performance.

\subsubsection{Named Entity Recognition}
We apply our \sys{} detection method to the NER task of TrojAI. The nature of NER is different from that of the SC task. SC only has one output per sample (i.e., whether the input sentence is a positive or negative movie review), while NER provides an output for every word in the input. This is because the NER task aims to classify each word by its type (i.e., what type of entity each word belongs to). 
TrojAI provides 80 input samples per NER model, while each input sample contains multiple words. Therefore, \sys{} detection method has enough samples to generate a reliable histogram with the given training samples.

Furthermore, we observe that the RAP trigger generates a large perturbation on the words adjacent to it compared to the one on other words residing in a different region of the sentence. 
To integrate this observation, we introduce the RAP trigger in a certain position of the sentence and only use the output of the words \textit{`adjacent'} to it. We move the RAP trigger to different positions of the input and collect the perturbations in different regions of the input.  
To control the range of the trigger's influence, we introduce an hyper-parameter $n\_adjacent$ that defines the maximal distance of the adjacent words to the RAP trigger.

TrojAI dataset~\cite{karra2020trojai} includes 54 trained BERT-based models for NER. These models use different tokenizers and architectures, which show different responses to the RAP trigger. We create a separate meta-classifier for each architecture setting. Each architecture has 18 models, of which 6 are Trojaned and the other 12 models are benign. We reserve 4 models (2 Trojaned and 2 benign) as the validation set and use the remaining 14 models (4 Trojaned and 10 benign) for meta-classifier training and hyper-parameter tuning (via repeated cross-validation).

Table~\ref{train_results_ner} shows the results when we use full optimization of the RAP trigger where we adjust the token embedding. Table~\ref{no_train_results_ner} shows the results when we use the relaxed RAP trigger variant of our method (i.e. we do not adjust the RAP trigger token embedding). 
Similar to the performance improvement observed on the SC task, we find out that using the relaxed RAP trigger achieves better detection results on the NER task compared to using the full RAP triggers.    
Also, as we discuss earlier, other benefits of \sys{}'s relaxed variant include shorted detection time and fewer hyper-parameters to tune. 

\begin{table}[!htb]
\centering
\scalebox{0.9}{
\begin{subtable}{0.35\textwidth}
\begin{tabular}{r|l|l|l}
 & Val & \begin{tabular}[c]{@{}l@{}}Cross\\ val\end{tabular} & Train\\ 
\hline
Acc & 100\% & 74.7\% & 84.6\% \\  
CE & 0.412 & 0.518 & 0.341 \\  
AUC & 100\% & 81.3\% & 100\% \\ 
\end{tabular}
\caption{\textit{roberta-base} architecture}
\end{subtable}
\begin{subtable}{0.35\textwidth}
\begin{tabular}{r|l|l|l}
 & Val & \begin{tabular}[c]{@{}l@{}}Cross\\ val\end{tabular} & Train\\ 
\hline
Acc & 75.0\% & 76.3\% & 92.9\% \\
CE & 0.744 & 0.604 & 0.335\\ 
AUC & 50.0\% & 85.8\% & 100\% \\ 
\end{tabular}
\caption{\textit{distilbert-base-cased} architecture}
\end{subtable}
\begin{subtable}{0.35\textwidth}
\begin{tabular}{r|l|l|l}
 & Val & \begin{tabular}[c]{@{}l@{}}Cross\\ val\end{tabular} & Train\\ 
\hline
Acc & 50.0\% & 57.5\% & 92.9\% \\ 
CE & 0.831 & 0.663 & 0.344 \\ 
AUC & 50.0\% & 65.8\% & 100\% \\ 
\end{tabular}
\caption{\textit{electra-small-discriminator} architecture}
\end{subtable}
}
\vspace{-0.5em}
\caption{Results on NER models from TrojAI dataset. RAP configuration: $n\_adjacent = 5$, $C\_low = 0.00007$, $C\_high = 0.00009$, $n\_epoch=50$, $lr=3000$. 
\label{train_results_ner}}
\end{table}

\begin{table}[!htb]
\centering
\scalebox{0.9}{
\begin{subtable}{0.35\textwidth}
\begin{tabular}{r|l|l|l}
 & Val & \begin{tabular}[c]{@{}l@{}}Cross\\ val\end{tabular} & Train\\ 
\hline
Acc & 100\% & 71.3\% & 84.6\% \\  
CE & 0.326 & 0.483 & 0.311  \\  
AUC & 100\% & 90.0\% & 100\% \\ 
\end{tabular}
\caption{\textit{roberta-base} architecture}
\end{subtable}
\begin{subtable}{0.35\textwidth}
\begin{tabular}{r|l|l|l}
 & Val & \begin{tabular}[c]{@{}l@{}}Cross\\ val\end{tabular} & Train\\ 
\hline
Acc & 75.0\% & 74.2\% & 100\% \\ 
CE & 0.596 & 0.620 & 0.275\\ 
AUC & 75.0\% & 80.8\% & 100\% \\ 
\end{tabular}
\caption{\textit{distilbert-base-cased} architecture}
\end{subtable}
\begin{subtable}{0.35\textwidth}
\begin{tabular}{r|l|l|l}
 & Val & \begin{tabular}[c]{@{}l@{}}Cross\\ val\end{tabular} & Train\\ 
\hline
Acc & 75.0\% & 68.5\% & 100\% \\ 
CE & 0.629 & 0.612 & 0.476 \\ 
AUC & 75.0\% & 62.5\% & 100\% \\ 
\end{tabular}
\caption{\textit{electra-small-discriminator} architecture}
\end{subtable}
}
\vspace{-0.5em}
\caption{Results on NER models from TrojAI dataset using relaxed RAP trigger. We use a sliding window of length $n\_adjacent = 5$. \label{no_train_results_ner}}
\end{table}

\subsubsection{Question Answering}

The question answering task also requires us to modify the implementation of \sys{} compared to the one of sentiment classification. 
For every word, QA outputs outputs its probability of being the beginning or end of the answer to the given question. Since most of the times the Trojan attacks make the prediction to be \textit{no answer} when the trigger is present in the input, we mainly focus on the output on the \textit{no answer} token (generally it is the [CLS] token at the beginning of the input).

\vspace{-0.5em}
\begin{table}[!htb]
\centering
\scalebox{0.9}{
\begin{subtable}{0.35\textwidth}
\begin{tabular}{r|l|l|l}
 & Val & \begin{tabular}[c]{@{}l@{}}Cross\\ val\end{tabular} & Train\\ 
\hline
Acc & 75.0\% & 58.1\% & 96.4\% \\  
CE & 0.553 & 0.590 & 0.306  \\  
AUC & 75.0\% & 72.5\% & 100\% \\ 
\end{tabular}
\caption{\textit{roberta-base} architecture}
\end{subtable}
\begin{subtable}{0.35\textwidth}
\begin{tabular}{r|l|l|l}
 & Val & \begin{tabular}[c]{@{}l@{}}Cross\\ val\end{tabular} & Train\\ 
\hline
Acc & 100\% & 100\% & 100\% \\ 
CE & 0.033 & 0.041 & 0.014\\ 
AUC & 100\% & 100\% & 100\% \\ 
\end{tabular}
\caption{\textit{distilbert-base-cased} architecture \label{no_train_results_qa_distilbert}}
\end{subtable}
\begin{subtable}{0.35\textwidth}
\begin{tabular}{r|l|l|l}
 & Val & \begin{tabular}[c]{@{}l@{}}Cross\\ val\end{tabular} & Train\\ 
\hline
Acc & 100\% & 91.3\% & 100\% \\ 
CE & 0.061 & 0.166 & 0.080 \\ 
AUC & 100\% & 98.5\% & 100\% \\ 
\end{tabular}
\caption{\textit{electra-small-discriminator} architecture \label{no_train_results_qa_electra}}
\end{subtable}
}
\vspace{-0.5em}
\caption{Results on QA models from the TrojAI dataset with relaxed RAP trigger and $5,000$ input samples from a public dataset (Squad\_v2~\cite{squad_v2}) to obtain model signatures. \label{no_train_results_qa}}
\end{table}

\begin{figure}[b!]
\centering
\includegraphics[width=0.9\textwidth]{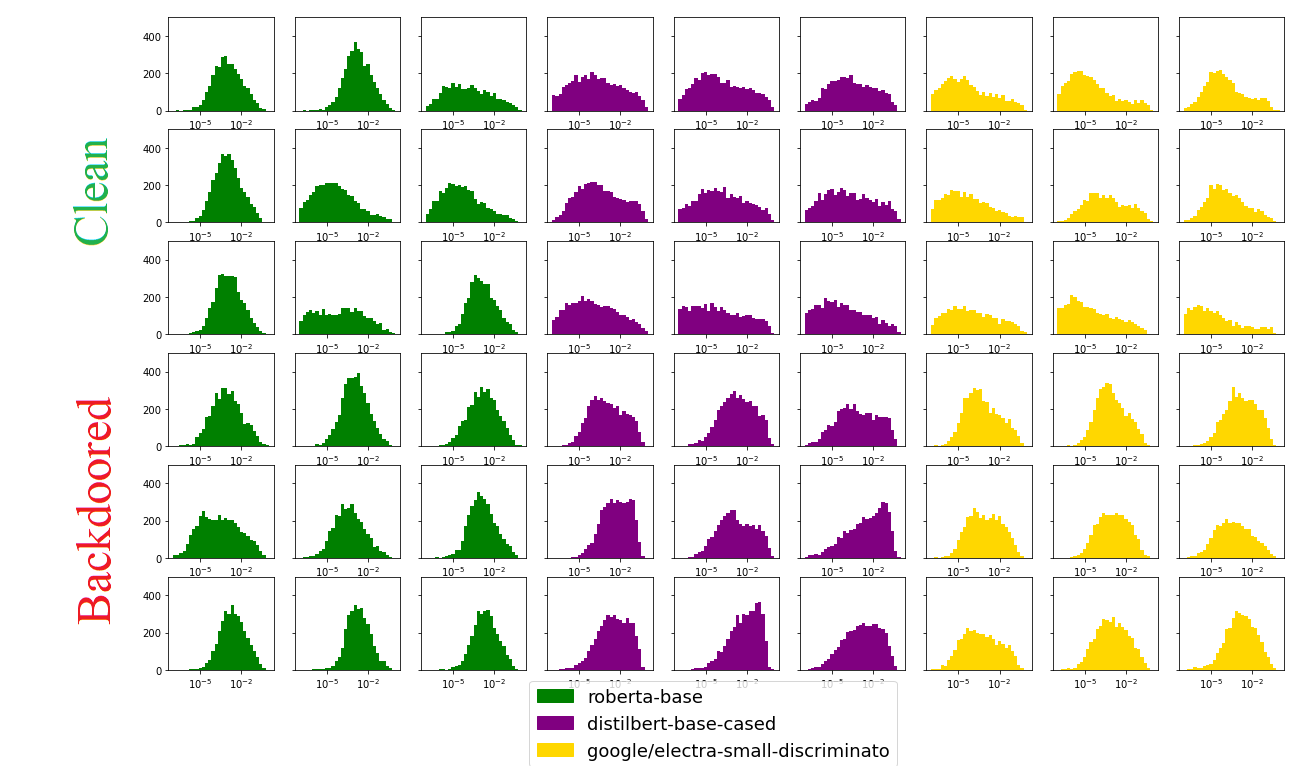}
\vspace{-0.6em}
\caption{The top 27 ‘signatures’ are from clean models, and the bottom 27 ‘signatures’ are from backdoored models. Each color represents a different model architecture. \label{image_qa_hist}}
\end{figure}

TrojAI dataset provides an average of 20 input samples for each QA model, which is not sufficient to generate a high-precision histogram. Therefore, we leverage \textbf{data augmentation} similar to our strategy in the SC task where we integrate data samples from a public dataset in the same domain. For the QA task, we use samples from the public dataset Squad\_v2~\cite{squad_v2}. 
Since we have empirically observed the effectiveness and efficiency of the relaxed RAP variant of \sys{} in both SC and NER tasks, we decide to use the relaxed RAP trigger for the QA task instead of the full RAP optimization.

The TrojAI \cite{karra2020trojai} dataset includes 96 BERT-based models trained for QA with different tokenizers and architectures, thus showing different responses to the RAP trigger. We create a separate meta-classifier for each architecture setting. Each architecture has 32 models, 20 of them Trojaned and the other 12 models benign. We reserve 4 models (2 Trojaned and 2 benign) as the validation set and use the remaining 28 models (18 Trojaned and 10 benign) to train the meta-classifier and run hyper-parameter tuning via repeated cross-validation.

We summarize the results of the QA task in Table~\ref{no_train_results_qa} and observe very promising results on the second and third architectures (subtables \ref{no_train_results_qa_distilbert}, \ref{no_train_results_qa_electra}). 
The signatures (i.e., histograms of the perturbations) of the QA models are visualized in Figure~\ref{image_qa_hist}.
We can see that with a logarithmic scale, the signatures of the Trojaned models and benign ones are visually different, which makes Trojan detection feasible. 
However, we cannot distinguish Trojan models and benign models from their signatures on the first architecture with all hyper-parameters we have tried.

\subsection{Ablation Study}

In this section, we investigate the impact of the design option on meta-classifier. Particularly, given the NLP task (e.g., SC, NER, or QA), we can decide if we train a \textit{separate} meta-classifier for a specific model architecture (e.g., roberta/distilbert/electra), or we train a \textit{unified} meta-classifier for different architectures. In the latter case, we merge the model histograms \sys{} extracts for all three model architectures and use the histograms combined to train the meta-classifier. 
We decide to design a different meta-classifier for each architecture on the SC task since the histograms seem to follow different distributions across architectures. 
However, as shown in Figure~\ref{image_qa_hist}, the histograms obtained for the QA task seem to have a smaller deviation across architecture, which inspires us to train a unified meta-classifier.

We compare the results with architecture-specific and architecture-agnostic meta-classifiers. Table~\ref{architecture-agnostic_sc}, Table~\ref{architecture-agnostic_ner}, and Table~\ref{architecture-agnostic_qa} show the results on SC, NER, and QA tasks respectively when using a \textbf{architecture-agnostic} meta-classifier. 
We use the same RAP hyper-parameters as the ones of the previously shown results for relaxed RAP. 
As a comparison, \sys{}'s detection results with \textbf{architecture-specific} meta-classifiers on SC, NER, and QA tasks are shown in tables~\ref{no_train_results}, Table~\ref{no_train_results_ner}, and Table~\ref{no_train_results_qa}, respectively. 

\begin{table}[!htb]
\centering
\begin{tabular}{r|l|l|l||l|l|l}
& \multicolumn{3}{c||}{architecture-agnostic} & \multicolumn{3}{c}{\begin{tabular}[c]{@{}l@{}}Average across \\ architecture-specific\end{tabular}} \\
 & Val & \begin{tabular}[c]{@{}l@{}}Cross\\ val\end{tabular} & Train & Val & \begin{tabular}[c]{@{}l@{}}Cross\\ val\end{tabular} & Train \\ 
\hline
Acc & 75\% & 69.2\% & 89.6\%    & 83.3\% & 69.2\% & 95.8\% \\ 
CE & 0.531 & 0.627 & 0.331      & 0.485 & 0.603 & 0.296 \\ 
AUC & 88.9\% & 72.9\% & 100\%   & 91.7\% & 72.9\% & 100\% \\ 
\end{tabular}
\caption{Comparison between architecture-agnostic meta-classifier and architecture-specific meta-classifiers on results on sentiment classification of TrojAI with relaxed RAP triggers and $1,000$ input samples from a public dataset (IMDB~\cite{imdb}) to obtain model signatures.  \label{architecture-agnostic_sc}}
\end{table}

\begin{table}[!htb]
\centering
\begin{tabular}{r|l|l|l||l|l|l}
& \multicolumn{3}{c||}{architecture-agnostic} & \multicolumn{3}{c}{\begin{tabular}[c]{@{}l@{}}Average across \\ architecture-specific\end{tabular}} \\
 & Val & \begin{tabular}[c]{@{}l@{}}Cross\\ val\end{tabular} & Train & Val & \begin{tabular}[c]{@{}l@{}}Cross\\ val\end{tabular} & Train \\ 
\hline
Acc & 50\% & 67.6\% & 95.1\%   & 83.3\% & 71.3\% & 94.9\% \\ 
CE & 0.664 & 0.623 &  0.291    & 0.517 & 0.572 & 0.354 \\ 
AUC & 63.9\% & 70.2\% & 100\%   & 83.3\% & 77.8\% & 100\% \\ 
\end{tabular}
\caption{Comparison between architecture-agnostic meta-classifier and architecture-specific meta-classifiers on results on NER models from TrojAI dataset using relaxed RAP trigger. We use a sliding window of length $n\_adjacent = 5$ \label{architecture-agnostic_ner}}
\end{table}

\begin{table}[!htb]
\centering
\begin{tabular}{r|l|l|l||l|l|l}
& \multicolumn{3}{c||}{architecture-agnostic} & \multicolumn{3}{c}{\begin{tabular}[c]{@{}l@{}}Average across \\ architecture-specific\end{tabular}} \\
 & Val & \begin{tabular}[c]{@{}l@{}}Cross\\ val\end{tabular} & Train & Val & \begin{tabular}[c]{@{}l@{}}Cross\\ val\end{tabular} & Train \\ 
\hline
Acc & 91.7\% & 88.8\% & 98.8\%   & 91.7\% & 83.1\% & 98.8\% \\ 
CE & 0.192 & 0.245 & 0.134      & 0.216 & 0.266 & 0.133 \\ 
AUC & 100\% & 97.6\% & 100\%    & 91.7\% & 90.3\% & 100\% \\ 
\end{tabular}
\caption{Comparison between architecture-agnostic meta-classifier and architecture-specific meta-classifiers on results on QA models from the TrojAI dataset with relaxed RAP trigger and $5,000$ input samples from a public dataset (Squad\_v2~\cite{squad_v2}) to obtain model signatures  \label{architecture-agnostic_qa}}
\end{table}

On the sentiment classification task, we can see  from Table~\ref{architecture-agnostic_sc} that using architecture-specific meta-classifiers achieves better results compared to a single architecture-agnostic, which is consistent with our intuition from the histogram distributions.   
On the named entity recognition task, we observe a clear performance decrement both on the validation data, and cross-validation evaluation from Table~\ref{architecture-agnostic_ner} when we use a single architecture-agnostic mete-classifier instead of training architecture-specific ones. 
Finally, on the question answering task, we see from Table~\ref{architecture-agnostic_qa} that using an architecture-agnostic mete-classifier gives better results.
This improvement is related to our observation that the histogram distributions of the QA models are similar across different architectures.

\section{Conclusion}

We develop a new model-level Trojan detection method for NLP applications using the change in the model's response (i.e. deviation of the confidence score) when we introduce a particular trigger to the input. We extract a histogram of the output perturbations generated from a set of inputs and use the histogram as the signature of the queried model. We train a meta-classifier to learn the signatures of benign and Trojaned models, thus producing the probability that the given model is Trojaned.
We demonstrate the effectiveness of our detection method across different NLP tasks (sentiment classification, named entity recognition, and question and answering) on public models from TrojAI \cite{karra2020trojai}. 
On sentiment classification, we also show results on a set of models that we created.

We observe that \textit{data augmentation} (incorporating input samples from public datasets) improves our detection performance. We corroborate this effect more in depth on the sentiment classification NLP task and compare the results with different levels of data augmentation.
Additionally, we propose a lightweight vatiant of \sys{} method with \textit{relaxed RAP triggers}. In this variant, we do not optimize the embedding of the RAP trigger, making the RAP trigger equivalent to a random noise added to the input.  
We empirically show the improved detection performance using the relaxed RAP method.
Another advantage of this variant is that the detection time is greatly reduced. Furthermore, not adjusting the RAP trigger embedding eliminates our need of accessing the model parameters. This means that we can treat the model as a black box and we only need oracle access to the queried model. 
Finally, we instigate when the defender shall use \textit{architecture-specific/agnostic meta-classifiers} depending on the histogram distributions obtained from the models with various architectures. If the deviation of the model histograms is large across different architectures, we find out that training separate meta-classifiers for each topology type gives better Trojan detection results. Otherwise, we observe that using a unified meta-classifier outperforms the architecture-specific one.

\bibliographystyle{ieeetr}
\bibliography{main} 

\end{document}